\def\year{2020}\relax

\documentclass[letterpaper]{article} 
\usepackage{aaai20}  
\usepackage{times}  
\usepackage{helvet} 
\usepackage{courier}  
\usepackage[hyphens]{url}  
\usepackage{graphicx} 

\usepackage{amsmath,amsfonts,amssymb}
\usepackage{stmaryrd}
\usepackage{xspace}
\usepackage[usenames,dvipsnames]{xcolor}
\usepackage{booktabs}
\usepackage{multirow}
\usepackage{subfig}
\usepackage{enumitem}
\usepackage{url}
\usepackage[labelfont=bf]{caption}
\usepackage[algoruled,boxed,linesnumbered]{algorithm2e}
\usepackage{todonotes}
\usepackage[capitalise,nameinlink]{cleveref}

\usepackage{graphicx}  
\title{Automated Discovery of Data Transformations for Robotic Process Automation}
\author{Volodymyr Leno\textsuperscript{1,2},
Marlon Dumas\textsuperscript{2},
Marcello La Rosa\textsuperscript{1},
Fabrizio M. Maggi\textsuperscript{2},
Artem Polyvyanyy\textsuperscript{1}\\
\textsuperscript{1}{University of Melbourne, Australia}\\
vleno@student.unimelb.edu.au,
artem.polyvyanyy@unimelb.edu.au, \\
marcello.larosa@unimelb.edu.au\\
\textsuperscript{2}{University of Tartu, Estonia}\\
marlon.dumas@ut.ee,
f.m.maggi@ut.ee}
 
\begin{document}

\maketitle

\begin{abstract}
Robotic Process Automation (RPA) is a technology for automating repetitive routines consisting of sequences of user interactions with one or more applications. In order to fully exploit the opportunities opened by RPA, companies need to discover which specific routines may be automated, and how. In this setting, this paper addresses the problem of analyzing User Interaction (UI) logs in order to discover routines where a user transfers data from one spreadsheet or (Web) form to another. The paper maps this problem to that of discovering data transformations by example -- a problem for which several techniques are available. The paper shows that a naive application of a state-of-the-art technique for data transformation discovery is computationally inefficient. Accordingly, the paper proposes two optimizations that take advantage of the information in the UI log and the fact that data transfers across applications typically involve copying alphabetic and numeric tokens separately. The proposed approach and its optimizations are evaluated using UI logs that replicate a real-life repetitive data transfer routine.
\end{abstract}

\section{Introduction}
\label{sec:intro}

Robotic Process Automation (RPA) is a technology that allows organizations to automate highly repetitive routines~\cite{DBLP:journals/misqe/LacityW16}. Among other things, RPA tools allow users to capture routines consisting of interactions with one or multiple applications (e.g., desktop and Web applications). These routines are encoded as scripts, which are then executed by software bots.


While RPA tools empower companies to automate routine work, they do not address the question of which specific routines should be automated, and how. Discovering all possible automatable routines is non-trivial, particularly in large organizations where work is highly distributed. This observation has triggered the development of techniques to analyze User Interface (UI) logs in order to discover repetitive routines that are amenable to automation via RPA tools~\cite{DBLP:conf/bpm/BoscoADRF19,DBLP:conf/caise/Ramirez19}.


One of the recurrent use cases for RPA bots is to automate data transfers across multiple applications, especially when these applications do not provide suitable APIs to enable their programmatic integration.
Starting from this observation, this paper addresses the problem of analyzing UI logs in order to discover routines where a user transfers data from one spreadsheet or (Web) form to another.


The paper addresses this problem by mapping it to that of discovering data transformations ``by example''. Indeed, the effect of transferring data from one spreadsheet or form to another is that one or more records (with an implicit schema) are mapped to one or more records with the same or a different schema. Hence, instead of discovering repetitive sequences of actions to transfer data across applications, we propose to discover the transformations that the users effectively perform when executing these sequences of actions.


The paper shows that a naive application of a state-of-the-art technique for data transformation discovery, namely Foofah~\cite{DBLP:conf/sigmod/JinACJ17}, is computationally inefficient. Accordingly, the paper proposes two optimizations that take advantage of the information in the UI log and the fact that data transfers across applications typically involve copying alphabetic and numeric tokens separately. The naive approach and its optimizations are evaluated using UI logs that replicate a real-life repetitive data transfer routine.

The next section introduces a motivating example. This is followed by an overview of RPA and related work on discovering automatable routines and data transformations (Sect.~\ref{sec:background}). Sect.~\ref{sec:approach} presents the approach and Sect.~\ref{sec:evaluation} its evaluation. Sect.~\ref{sec:conclusion} summarizes the contribution and future work.

\section{Motivating Example}
\label{sec:running}

\begin{figure}[t]
    \centering
    \includegraphics[scale = 0.33]{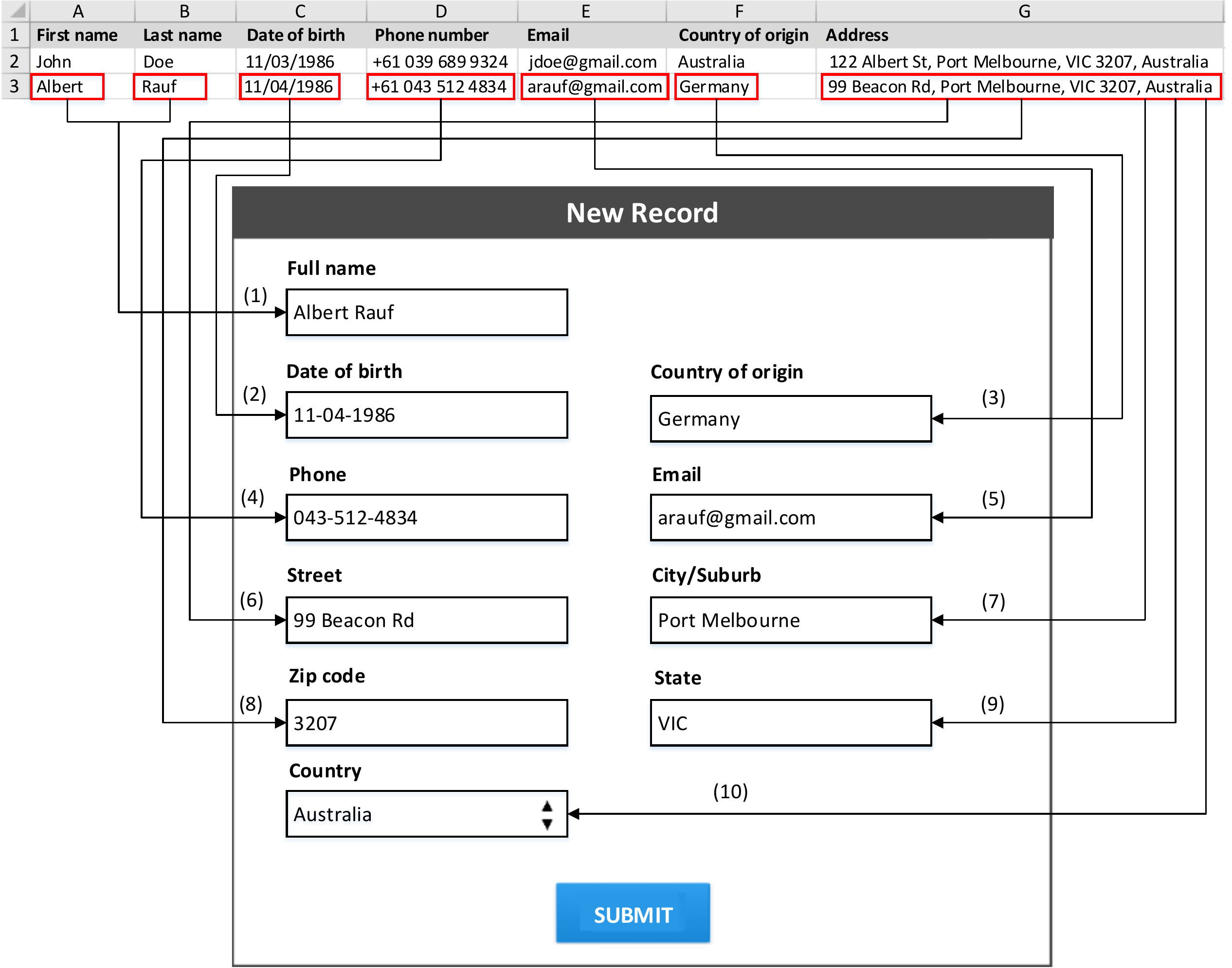}
    \caption{Motivating example: excerpt of spreadsheet (top), excerpt of Web-form (bottom).}
    \label{fig:example}
		\vspace{-5mm}
\end{figure}

Below, we describe an RPA use case inspired by a real-life scenario at the University of Melbourne.
The scenario concerns the transfer of student records (e.g., name, surname, phone, email, address) from an Excel spreadsheet to the data-entry Web-form of a students management system, as part of the student admission process at the university. \figurename~\ref{fig:example} illustrates this data transferring task. For each row in the spreadsheet (representing a student), a student admission staff manually copies every cell in that row and pastes that into the corresponding text field in the Web-form. Once the data transfer for a given student has been completed, the staff presses the Submit button to submit the data into the students management system, and repeats the procedure for the next row. Such data transferring tasks often involve data transformations as source and target data can be in a different format. The user has to copy data from a certain cell, paste it into the corresponding field and do some manual manipulations to convert it into the required format. For each data element of the target, \figurename~\ref{fig:example} shows the mapping to the corresponding data elements from the source. Such manipulations can be automated via an RPA bot so long as they are described via a transformation program.

Depending on the number of source and target elements in a transformation, transformations can be of four different types. \textit{One-to-one (1 - 1)} transformations involve one source and one target element. Usually, they are represented by manipulations that correspond to characters replacement (\#2 in Fig.~\ref{fig:example}) or partial copying (\#4). When the data is already in the target format, no manipulations are required and the user has to perform a simple copy-paste operation (\#3, \#5). The next transformation type is \textit{one-to-many (1 - N)}. This type usually describes manipulations that require a split operation. In the example of \figurename~\ref{fig:example}, the user has to split the full address from the spreadsheet into five different elements: street, city, ZIP code, region and country (\#6-\#10). Next, \textit{many-to-one (N - 1)} transformation require a merge operation over several elements. In our example, cells from the spreadsheet that contain the student's first and last name have to be joined to fill in the text field that corresponds to the full name in the Web form (\#1). Finally, \textit{many-to-many (N - N)} transformations involve manipulations such as copying and pasting a range of cells, pivoting a row into a column in a spreadsheet, etc. These are not shown in \figurename~\ref{fig:example}.

\section{Background and Related Work}
\label{sec:background}

In this section, we give an overview of RPA, we introduce the concept of UI log and its structure, and we present existing techniques for transformation discovery.

\subsection{Robotic Process Automation}
\label{sec:rpa}

RPA is a technology that aims at automating human tasks by means of software robots (bots for short) by mimicking the actions performed by users across various applications during the execution of a task. RPA may lead to improvements in efficiency and data quality in business processes involving clerical work \cite{DBLP:conf/wea2/AguirreR17}.
Nowadays, there are many commercial RPA solutions such as Automation Anywhere Enterprise RPA \footnote{\url{https://www.uipath.com/}} and UiPath Enterprise RPA Platform \footnote{\url{https://www.auomationanywhere.com/}}, which provide capabilities to create scripts both via record-and-replay approaches or via manual coding.

Two broad types of RPA use cases are generally distinguished: attended and unattended. In attended automation, the bot assists a user in performing her daily tasks. During its execution, the bot can take input data from a user, it can be interrupted, paused or stopped at any time. Attended bots usually run on local machines and manipulate the same applications as a user. They can be used to automate routines that require dynamic inputs, human judgement or when the routine is likely to have exceptions. Meanwhile, in unattended automation, the bot works independently, without user involvement. Unattended bots are used for back-office functions. They usually run on organization's servers and are suitable for executing deterministic routines where all inputs and outputs are known in advance, and all executions paths are well defined.

\subsection{UI log}
\label{sec:uiLog}

A UI log is a chronologically ordered sequence of actions performed by a user in a single workstation during her interactions with different applications when performing a task. Each row represents a certain action and every action is characterized by a timestamp (when the action was performed), a type (what action was performed) and a source (where the action was performed). It also contains other information that describes the application elements that were involved (e.g., label of a button, url, address of a cell and its value, etc.), called \textit{payload}. \tablename~\ref{tab:recording} shows the extract of a UI log that was recorded during the execution of the data transferring task illustrated in Section \ref{sec:running}.

\begin{table*}[h!]
\caption{Example of UI log. \label{tab:recording}}
\centering
\scalebox{0.63}{
\begin{tabular}{| c | l |  l | l | l | l | l |}
\hline
 & \textbf{Timestamp} & \textbf{Action Type} & \textbf{Source} & \textbf{Content} & \textbf{Field name} & \textbf{Field value} \\ \hline
1 & 2019-03-03T19:02:18 & Copy cell & Worksheet & ``Albert" & A3 & ``Albert" \\
2 & 2019-03-03T19:02:23 & Click field & Web &  & `Full Name & ``" \\
3  & 2019-03-03T19:02:26 & Paste & Web & ``Albert" & Full Name & ``" \\
4 & 2019-03-03T19:02:27 & Edit field & Web & & Full Name & ``Albert" \\
5 & 2019-03-03T19:02:32 & Copy cell & Worksheet & ``Rauf" & B3 & ``Rauf" \\
6 & 2019-03-03T19:02:35 & Click field & Web & & Full Name & ``Albert" \\
7  & 2019-03-03T19:02:37 & Paste & Web & ``Rauf" & Full Name & ``Albert" \\
8 & 2019-03-03T19:02:39 & Edit field & Web & & Full Name & ``Albert Rauf" \\

9 & 2019-03-03T19:02:43 & Copy cell & Worksheet & ``Germany" & F3 & ``Germany" \\
10 & 2019-03-03T19:02:45 & Click field & Web & & Country & ``" \\
11  & 2019-03-03T19:02:46 & Paste & Web & ``Germany" & Country & ``" \\
12 & 2019-03-03T19:02:47 & Edit field & Web & & Country & ``Germany" \\

13 & 2019-03-03T19:02:50 & Copy cell & Worksheet & ``11/04/1986" & C3 & ``11/04/1986" \\
14 & 2019-03-03T19:02:52 & Click field & Web & & Date & ``" \\
15 & 2019-03-03T19:02:53 & Paste & Web & ``11/04/1986" & Date & ``" \\
16 & 2019-03-03T19:02:58 & Edit field & Web & & Date & ``11-04-1986" \\

17 & 2019-03-03T19:03:01 & Copy cell & Worksheet & ``+ 61 043 512 4834"  & D3 & ``+ 61 043 512 4834" \\
18 & 2019-03-03T19:03:05 & Click field & Web & & Phone & ``" \\
19 & 2019-03-03T19:03:07 & Paste & Web & ``+ 61 043 512 4834"  & Phone & ``" \\
20 & 2019-03-03T19:03:13 & Edit field & Web & & Phone & ``043-512-4834"  \\

21 & 2019-03-03T19:03:18 & Copy cell & Worksheet & ``arauf@gmail.com" & E3 & ``arauf@gmail.com" \\
22 & 2019-03-03T19:03:21 & Click field & Web & & Email & ``" \\
23  & 2019-03-03T19:03:23 & Paste & Web & ``arauf@gmail.com" & Email & ``" \\
24 & 2019-03-03T19:03:24 & Edit field & Web & & Email & ``arauf@gmail.com" \\

25 & 2019-03-03T19:03:30 & Copy cell & Worksheet & ``99 Beacon Rd, Port Melbourne, VIC 3207, Australia" & G3 & ``99 Beacon Rd, Port Melbourne VIC 3207, Australia" \\
26 & 2019-03-03T19:03:33 & Click field & Web & & Adress\_Street & ``" \\
27 & 2019-03-03T19:03:36 & Paste & Web & ``99 Beacon Rd, Port Melbourne, VIC 3207, Australia" & Adress\_Street & ``" \\
28 & 2019-03-03T19:03:43 & Edit field & Web & & Adress\_Street & ``99 Beacon Rd" \\

29 & 2019-03-03T19:03:48 & Click field & Web & & Address\_City & ``" \\
30  & 2019-03-03T19:03:51 & Paste & Web & ``99 Beacon Rd, Port Melbourne, VIC 3207, Australia" & Address\_City & ``"\\
31 & 2019-03-03T19:03:56 & Edit field & Web & & Address\_City & ``Port Melbourne" \\

32 & 2019-03-03T19:04:00 & Click field & Web & & Adress\_Region & ``" \\
33 & 2019-03-03T19:04:02 & Paste & Web & ``99 Beacon Rd, Port Melbourne, VIC 3207, Australia" & Address\_Region & ``" \\
34 & 2019-03-03T19:04:10 & Edit field & Web & & Address\_Region & ``VIC" \\

35 & 2019-03-03T19:04:13 & Click field & Web & & Address\_ZipCode & ``" \\
36 & 2019-03-03T19:04:15 & Paste & Web & ``99 Beacon Rd, Port Melbourne, VIC 3207, Australia" & Address\_ZipCode & ``" \\
37 & 2019-03-03T19:04:22 & Edit field & Web & & Address\_ZipCode & ``3207" \\

38 & 2019-03-03T19:04:25 & Click field & Web & & Address\_Country & ``" \\
39 & 2019-03-03T19:04:29 & Edit field & Web & & Address\_Country & ``Australia" \\

40 & 2019-03-03T19:04:34 & Click check box & Web & & International & ``FALSE" \\
41 & 2019-03-03T19:04:34 & Edit field & Web & & International & ``TRUE" \\

42 & 2019-03-03T19:04:45 & Click button & Web & & Submit & \\

\hline
\end{tabular}
}
\label{table:recording}
\vspace{-3mm}
\end{table*}

A UI log can describe multiple executions of a task. One execution of a task is called \textit{task trace} and it contains a sequence of actions required to complete the task.

\subsection{Related Work}
\label{sec:related}

Discovering RPA routines is related to the problem of Automated Process Discovery (APD) \cite{DBLP:journals/tkde/AugustoCDRMMMS19}, which has been studied in the field of process mining. Recent work~\cite{DBLP:conf/bpm/Geyer-Klingeberg18,DBLP:conf/caise/Ramirez19} shows that RPA can benefit from process mining. In particular, the work in \cite{DBLP:conf/caise/Ramirez19} proposes to apply traditional APD techniques to discover process models of routines captured in UI logs. However, traditional APD techniques discover control-flow models, while, in the context of RPA, we seek to discover executable specifications that capture the mapping between the outputs and the inputs of the actions performed during a routine.




Discovering RPA routines is also related to the problem of Web form and table auto-completion. Excel's Flash Fill feature, for example, detects string patterns in the values of the cells in a spreadsheet and uses these patterns for auto-completion~
\cite{DBLP:conf/popl/Gulwani11}. This and similar approaches focus on partial automation. They do not seek to discover executable specifications of entire data transfer routines.


A closely related proposal is that of~\cite{DBLP:conf/bpm/BoscoADRF19}, which proposes a technique to discover sequences of actions such that the inputs of each action in the sequence (except the first one) can be derived from the data observed in previous actions. This latter technique can only discover perfectly sequential routines, and is hence not resilient to variability in the order of the actions, whereas in reality, different users may perform the actions in a routine in a different order.
The approach presented in \cite{DBLP:conf/otm/GaoZLA19} aims at extracting rules from UI logs that can be used to automatically fill in forms. This approach, however, does not discover any data transformations (i.e., relations between outputs and inputs).

The problem of discovering data transformations from examples is well studied, particularly in the context of data wrangling. The authors in \cite{DBLP:conf/pldi/BarowyGHZ15} propose an approach to extract a structured relational database from semi-structured spreadsheets. Kandel et. al \cite{DBLP:conf/chi/KandelPHH11} present an interactive system that suggests a list of syntactic manipulation operations based on given examples of inputs. However, only a limited amount of operations are supported and considering that the user provides only input examples, the list of suggested manipulations is too large as it is hard to predict the desired transformation without knowing the target format. Data preparation tools such as Trifacta\footnote{\url{https://www.trifacta.com/}} and OpenRefine\footnote{\url{http://openrefine.org/}} allow users to specify transformations by writing a program in a special purpose language. These approaches, however, aim at assisting the user when transforming the data into the target format, rather than discovering the underlined transformations.


In this paper, we use a data transformation-by-example discovery technique, called Foofah \cite{DBLP:conf/sigmod/JinACJ17}. Given a set of input-output examples this technique aims at synthesizing an optimal transformation program that describes what manipulations have to be performed to convert raw input data into the required format. Foofah describes program synthesis as a search problem in a state space graph, where each edge represents a modification operation (e.g., split, removal, etc.), start and end nodes describe raw and final data, and all other nodes denote an intermediate value during the transformation process. It then uses a heuristic search approach based on the A* algorithm to find an optimal path from the initial state to the goal state. The cost function is defined as the minimum number of manipulations required to transform one value into another. We selected Foofah because, unlike other transformation discovery approaches, it can handle both string and tabular transformations.

\section{Approach}
\label{sec:approach}

\urldef{\footurla}\url{https://github.com/apromore/RPA_SemFilter/releases}

\begin{figure*}[t]
		\vspace{-2mm}
    \centering
    \includegraphics[width=0.5\textwidth]{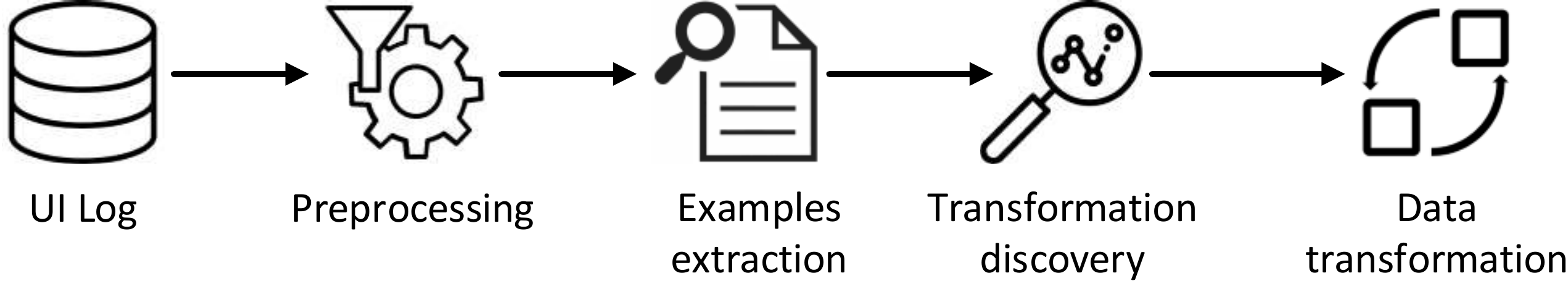}
    \caption{Baseline approach for discovering data transformations in UI logs.}
    \label{fig:approach}
		\vspace{-3mm}
\end{figure*}

Our baseline approach for discovering data transformations from UI logs is summarized in \cref{fig:approach}.
Once a UI log is recorded, it is preprocessed.
Then, a collection of transformation examples is extracted from the preprocessed log, followed by the discovery of data transformations.
This baseline approach is described in \cref{sec:preprocessing,sec:naive}.
We then present two optimizations of the baseline approach.
The first optimization aims at reducing the number of fields in the transformation examples (\cref{sec:opt1}) while the second optimization (\cref{sec:opt2}) reduces the number of transformation examples given as input to the discovery step.

\subsection{Preprocessing}
\label{sec:preprocessing}

A UI log is a chronologically ordered sequence of actions performed by a user during interactions with various applications (e.g., Desktop and Web-based applications).
Usually, this sequence describes multiple executions of the same task (i.e., several \emph{task traces}).
The first step of preprocessing is to identify task traces.
The process of identifying task traces is called \textit{segmentation}.
Segmentation can be accomplished in several ways, for example using sessionization techniques from the field of Web session mining~\cite{DBLP:journals/corr/abs-1004-1257}.
As segmentation is not in the focus of this work, we assume that all task traces have the same (known) end action (e.g., pressing the submit button in the Web form).
With this assumption, it is trivial to split the UI log into task traces.

A UI log may contain redundant actions that do not affect the outcome of a task (e.g., copying without pasting and navigating across cells in a spreadsheet without copying their content).
Such actions must be removed.
Therefore, a sub-step of preprocessing is \textit{filtering}.
We developed an auxiliary tool that filters out redundant actions in a semantics-preserving manner.\footnote{\footurla}
The tool works by applying regular expression find-and-replace operations.
It uses a collection of predefined rules that capture redundant action patterns and how to filter them out (e.g., when there are two consecutive copy actions without a paste action in between, the first copy action can be ignored).
Some predefined rules target the control-flow perspective (e.g., navigation and subsequent copy actions without pasting in between).
Other rules are data-aware (e.g., double editing of a text field with replacement).
The filtering is applied to each task trace. The output is a collection of task traces without redundant actions.

\subsection{Baseline Approach}
\label{sec:naive}

Given a collection of filtered task traces, the first step of our baseline approach for data transformation discovery is to extract \textit{transformation examples}.
A transformation example is a tuple (\textit{I}, \textit{O}, \textit{S}, \textit{T}), where \textit{I} is the raw data from a source document \textit{S}, and \textit{O} is the data in a final format stored in a target document \textit{T}.
A document is an instance of an application that was used to accomplish a task (e.g., a spreadsheet or Web form).
For each task trace in a UI log, we extract all the values used in the source document \textit{S} and the target document \textit{T}.
All these values are then used to build a transformation example.
Considering our running example, one collection of target values is:

\smallskip
\emph{O = [``Albert Rauf'',``11-04-1986'', ``Germany'', ``043-512-4834'', ``arauf@gmail.com'', ``99 Beacon Rd'', ``Port Melbourne'', ``VIC'', ``3207'', ``Australia'']},

\smallskip
\noindent
while the corresponding collection of source values is:

\smallskip
\emph{I = [``Albert'', ``Rauf'', ``11/04/1986'', ``+61 043 512 4834'', ``arauf@gmail.com'', ``Germany'', ``99 Beacon Rd, Port Melbourne, VIC 3207, Australia'']}.

\smallskip

After all transformation examples are extracted, they are provided as input to a transformation discovery technique.
In this work, we use a data transformation-by-example discovery technique, called Foofah~\cite{DBLP:conf/sigmod/JinACJ17}.
Given transformation examples \textcolor{black}{summarized in the form of input and output tables}, this technique aims at synthesizing an optimal transformation program that describes what manipulations have to be performed to convert raw input data into the required final format.\footnote{For details on Foofah's syntax and implementation, we refer the reader to \cite{DBLP:conf/sigmod/JinACJ17}} \textcolor{black}{Foofah supports the Potter's Wheel operations~\cite{DBLP:conf/vldb/RamanH01} and is extensible.
We assume that the output is noise- and error-free, meaning that the analyzed data transformations are correct.}

\begin{figure}[h]
\centering
\includegraphics[scale = 0.56]{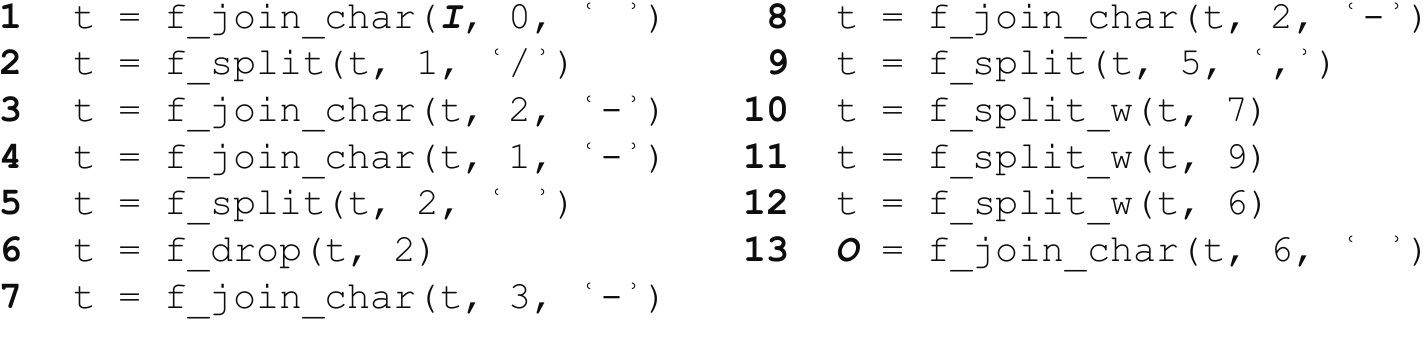}
\caption{A sample transformation program (pseudocode).}
\label{fig:baselineTransformation}
\vspace{-2mm}
\end{figure}

Nonetheless, Foofah may face scalability limitations in the case when the target transformation is not trivial (i.e., cannot be described using a single join or split command).
In addition, it can take a long time to discover a transformation based on a large number of examples, and, in some cases, Foofah may fail to discover a transformation program, even if such exists.
For example, Foofah executed for 310 seconds failed to discover a transformation program for the example described above.
However, such a transformation program indeed exists.
\cref{fig:baselineTransformation} presents a sample program that implements the transformation.
The program takes \textcolor{black}{table} $I$ as input and produces \textcolor{black}{table} $O$ as output.
The intermediate results of operations are stored in variable $t$.
The sample transformation program in \cref{fig:baselineTransformation} consists of join and split commands.
For instance, the \texttt{f\_join\_char} command at line~1 merges first name and last name into full name and places one space character (cf. the third parameter) between them.
The second parameter 0 tells that the element at the first position in table $I$ (\textit{``Albert''}) should be joined with its immediate successor (\textit{``Rauf''}); note that we count positions from zero and join two subsequent elements.
The \texttt{f\_split} command at line~2 divides the date into day, month, and year components using the `/' symbol;
note that the command takes the result of line~1, i.e., table \texttt{t}, as input, which in the second position, referenced by the value 1 of the second parameter, contains the string \textit{``11/04/1986''}.

%

\textcolor{black}{The time complexity of Foofah technique is $O((kmn)^d)$, where \textit{m} and \textit{n} are, respectively, the number of cells in the input and output tables, \textit{k} is the number of candidate data operations for each intermediate transformation result, and \textit{d} is the number of components in the synthesized program~\cite{DBLP:conf/sigmod/JinACJ17}.
The number of cells is $\mathit{cols} \times \mathit{rows}$, where \textit{cols} is the number of columns and \textit{rows} is the number of rows in the table.
For the baseline approach, \textit{cols} corresponds to the number of data fields, and \textit{rows} represents the number of transformation examples given as input to Foofah.}

Next, we present two improvements to the baseline approach that aim at reducing both the number of data fields and transformation examples that are provided to Foofah for processing, thus improving the overall performance.

\subsection{Grouping Examples by Targets}
\label{sec:opt1}

Document-to-document transformations usually involve complex manipulations.
Thus, it takes a significant amount of time to synthesize a corresponding transformation program, and, often, Foofah fails to discover any transformation.
We propose an optimization that aims at decomposing the transformation from document-to-document to source-to-target level.
This is done by projecting transformation examples onto target elements within a target document \textcolor{black}{using the information about data elements involved in task executions available in the UI log}.
For instance, the transformation example with input table  $I$
and output table $O$ from \cref{sec:naive}
projected onto the target text field \textit{Full name} results in the transformation example with input $I'$ = \textit{[``Albert'', ``Rauf'']} and output $O'$ = \textit{[``Albert Rauf'']}.
For the obtained transformation example, it is easy to find the corresponding transformation of join by space operation.

Considering this optimization, the approach is the following.
For each task trace, we extract the last edit of the elements within the target application.
For each such edit, we identify its type (e.g., whether the user pasted into the field or filled it in manually).
If the user pasted into the field (edit action is preceded by paste action into the same field), the sources that contributed to the final value of the target field are identified.
For example, the final value ``Albert Rauf'' of the \textit{Full name} text field in the Web form was obtained from the concatenation of A3 (``Albert'') and B3 (``Rauf'') cells in the spreadsheet, refer to \cref{fig:example}.
For each paste action in the field, we find the most recent copy action that contains information about the source and its value that was copied.
Together with the target value, this information is used to build a transformation example.
\cref{alg:examplesExtraction} shows the procedure for extracting transformation examples.
In the case of manual input, it is not possible to trace back the source, and, therefore, the corresponding edit is ignored.

\vspace{-2mm}
\begin{algorithm}[h]
\SetAlgoLined
\SetKwInOut{Input}{input}\SetKwInOut{Output}{output}
\Input{UI log $L$}
\Output{Input-output examples $\mathit{IO}$}
$F \leftarrow \mathit{extractTargetFields(L)}$\;
$\mathit{IO} \leftarrow \emptyset$\;
\ForEach{target field $f \in F$}{
	$\mathit{source} \leftarrow \emptyset$\;
	$\mathit{input} \leftarrow \emptyset$\;
	$\mathit{output} \leftarrow \mathit{valueOfLastEdit(f)}$\;
	\ForEach{paste $p$ in $f$}{
		$\mathit{copy} \leftarrow \mathit{lastCopy(p)}$\;
		$\mathit{source} \leftarrow \mathit{source} \cup \mathit{getSource(copy)}$\;
		$\mathit{input} \leftarrow \mathit{input} \cup \mathit{getValue(copy)}$\;
	}
	$\mathit{IO} \leftarrow \mathit{IO} \cup \{(input, output, source, f)\}$
}
\caption{Transformation examples extraction}\label{alg:examplesExtraction}
\end{algorithm}
\vspace{-2mm}

\begin{figure}[t]
    \centering
    \includegraphics[scale = 0.64]{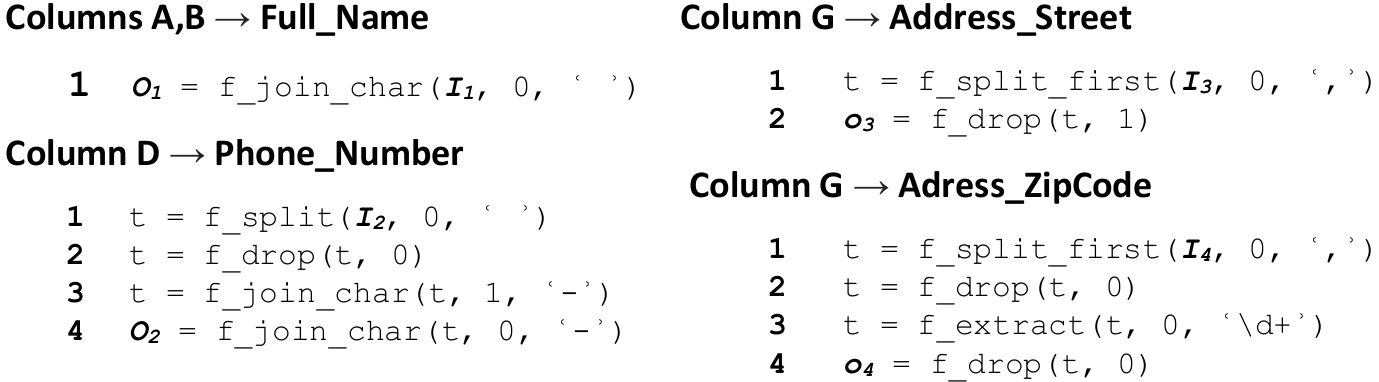}
    \caption{Transformation programs synthesized by the approach that groups transformation examples by targets.}
    \label{fig:opt1Transformation}
		\vspace{-5mm}
\end{figure}

After obtaining a set of transformation examples, the approach splits it into smaller groups based on targets.
For each obtained group, we then discover a separate transformation program using Foofah.
\cref{fig:opt1Transformation} presents the transformations discovered using this approach.


\subsection{Grouping Examples by Input Structure}
\label{sec:opt2}

The first optimization seeks to reduce the number of input fields that Foofah needs to consider, but it does not reduce the number of transformation examples given as input.
Thus, when a complex transformation occurs between one or multiple source elements and one target element, Foofah may still fail to synthesize a transformation program.
Accordingly, the number of transformation examples used to synthesize the transformation program has to be reduced.
On the other hand, to discover a correct transformation program, a sufficient amount of examples that capture different cases and behaviors is required.
Therefore, it is crucial to preserve all the behavior during the reduction process.

There are further limitations of Foofah that are not addressed by the improvement presented in \cref{sec:opt1}.
For example, Foofah does not discover conditional transformations, where different manipulations are applied depending on the input.
Consequently, it cannot deal with heterogeneous data.
When the data comes from different sources, it can be stored in different formats, and Foofah will fail to discover a transformation program.
Another limitation of Foofah is that it can discover transformations only if the output is not ambiguous, i.e., it is clear from which components it was obtained.
Consider the example in \tablename~\ref{tab:ambiguousExample}, where the user aims to extract a ZIP code from a full address.
While for Example 3 it is clear that ZIP code ``3205'' was obtained from Address 3 (``396 Clarendon St, South Melbourne, VIC 3205, Australia''), the origin of ZIP code ``3207'' is ambiguous as it can be obtained from Address 1 (``122 Albert St, Port Melbourne, VIC 3207, Australia'') or Address 2 (``99 Beacon Rd, Port Melbourne, VIC 3207, Australia''). In this case, Foofah will not discover any transformation.

\begin{table}[h!]
\caption{Transformation example with ambiguous output.
\label{tab:ambiguousExample}}
\vspace{-2mm}
\centering
\scalebox{0.8}{
\begin{tabular}{| c | l | c |}
\hline
\textbf{No.} & \textbf{Input} & \textbf{Output} \\
\hline
\hline
1 & 122 Albert St, Port Melbourne,  & 3207\\
& VIC 3207, Australia & \\
\hline
2 & 99 Beacon Rd, Port Melbourne, & 3207 \\
& VIC 3207, Australia & \\
\hline
3 & 396 {Clarendon St, South Melbourne, } & 3205 \\
& VIC 3205, Australia & \\
\hline
\end{tabular}
}
\end{table}

To address these limitations, we group the target examples into equivalence classes, where each class represents a different structural pattern of the input data.
For the input data of each transformation example, we discover the corresponding symbolic representation that describes its structural pattern.
This is achieved by applying \textit{tokenization}.
During tokenization, each maximal chained subsequence of symbols of the same type (e.g., digits or alphabetic characters) is replaced with a special token character, while special symbols (e.g., separators and punctuation marks) remain unchanged.
A maximal subsequence of digits is replaced with the \textless d\textgreater+ symbol.
A maximal subsequence of alphabetic characters is replaced with the \textless a\textgreater+ token.
Below, we show how a raw address value from our running example is tokenized:

\medskip
\noindent
\small s = ``99 Beacon Rd, Port Melbourne, VIC 3207, Australia''
\normalsize

\smallskip
\noindent
1) Replace all alphabetic characters with tokens:

\smallskip
\noindent
\small s = ``99 \textless a\textgreater+ \textless a\textgreater+, \textless a\textgreater+ \textless a\textgreater+, \textless a\textgreater+ 3207, \textless a\textgreater+''
\normalsize

\smallskip
\noindent
2) Replace all digits with tokens:

\smallskip
\noindent
\small s = ``\textless d\textgreater+ \textless a\textgreater+ \textless a\textgreater+, \textless a\textgreater+ \textless a\textgreater+, \textless a\textgreater+ \textless d\textgreater+, \textless a\textgreater+''

\smallskip
\normalsize
\noindent
Hence, the resulting pattern is
``\textless d\textgreater+ \textless a\textgreater+ \textless a\textgreater+, \textless a\textgreater+ \textless a\textgreater+, \textless a\textgreater+ \textless d\textgreater+, \textless a\textgreater+''.
Note that all the space characters as well as punctuation are preserved in the pattern.

\begin{figure}[t]
    \centering
    \includegraphics[scale = 0.65]{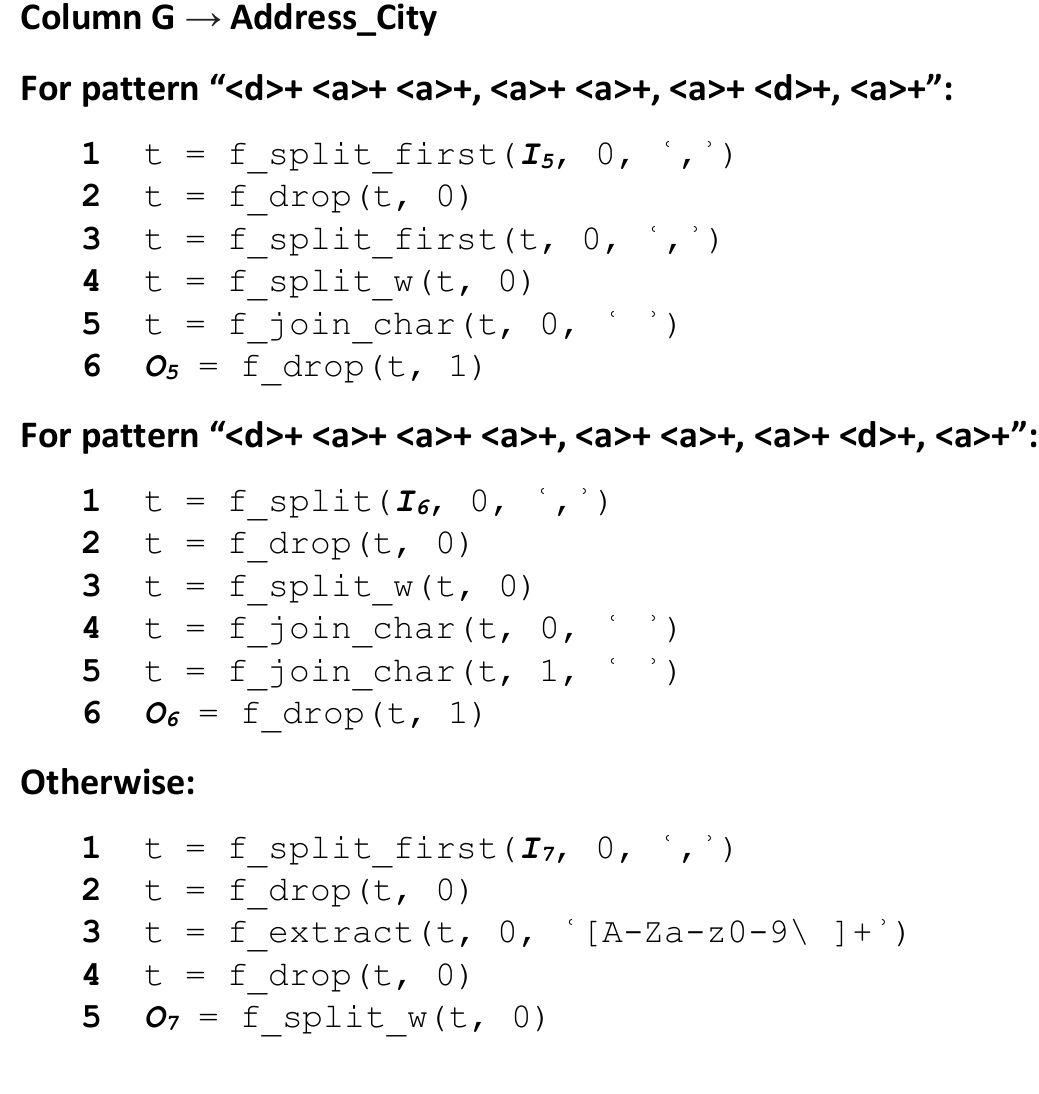}
    \caption{Transformation discovered by approach that combines both optimizations.}
    \label{fig:opt2Transformation}
		\vspace{-5mm}
\end{figure}

\textcolor{black}{After tokenization, we group the transformation examples by structural patterns, i.e., we create one group of examples per pattern. For each group, we discover a transformation program by providing to Foofah one randomly selected transformation example from the group. Providing multiple examples (or all examples) is useful when the examples are heterogeneous. If the examples are homogeneous, which is what the grouping step ensures, Foofah is more likely to synthesize a transformation program from a single example.\footnote{It may happen that the synthesized transformation does not generalize to all examples in a group, e.g., due to ambiguity in the mapping from inputs to outputs. Hence, we check that the transformation discovered from one example fits all examples in the group. If it does not, we try to discover a transformation from all examples in the group. If the latter fails, the discovery procedure fails.}
Sometimes, several groups (i.e., patterns) may lead to the same transformation program. If this occurs, the corresponding groups are merged.
} The transformation programs synthesized by Foofah for our running example using this optimization are shown in \cref{fig:opt2Transformation}.


\begin{table*}[t!]
\caption{Transformation discovery results.\label{tab:transformationDiscoveryResults}}
\centering
\scalebox{0.82}{
\begin{tabular}{| c | c | c | c | c |}
\hline
\textbf{Transformation} & \textbf{Example} & \textbf{Baseline} & \textbf{Opt 1} & \textbf{Opt 1 + Opt 2} \\
\textbf{type} & & & & \\ \hline
N - 1 & ``Igor", ``Honchar" $\Rightarrow$ \ ``Igor Honchar" & \textbf{1.295} & 1.584 & 1.745 \\ \hline
1 - 1 & ``18/08/1992" $\Rightarrow$ \ ``18-08-1992" & 6.584 & 6.639 & \textbf{0.476} \\
1 - 1 & ``+61 029 211 4904" $\Rightarrow$ ``029-211-4904" & N/A (2306.036) & N/A (2271.19) & \textbf{0.5086} \\
1 - 1 & ``New Zealand"  $\Rightarrow$ ``New Zealand" & \textbf{0.347} & 0.392 & 0.704 \\
1 - 1 & ``wmacdonald@gmail.com" $\Rightarrow$ ``wmacdonald@gmail.com" & \textbf{0.34} & 0.391 & 0.397 \\ \hline
1 - N & ``122 Albert St, Port Melbourne, VIC 3207, Australia" $\Rightarrow$ & timeout & 7504.934 & \textbf{85.423} \\
& ``122 Albert St", ``Port Melbourne", ``VIC", ``3207" & & & \\ \hline
1 - 1& ``122 Albert St, Port Melbourne, VIC 3207, Australia" $\Rightarrow$ ``122 Albert St" & - & \textbf{1.243} & 1.55 \\
1 - 1& ``122 Albert St, Port Melbourne, VIC 3207, Australia" $\Rightarrow$ ``Port Melbourne" & - & N/A (1983.501) & \textbf{54.777} \\
1 - 1 & ``122 Albert St, Port Melbourne, VIC 3207, Australia" $\Rightarrow$ ``VIC" & - & timeout & \textbf{26.603} \\
1 - 1& ``122 Albert St, Port Melbourne, VIC 3207, Australia" $\Rightarrow$ ``3207" & - & N/A (1884.397) & \textbf{2.49} \\ \hline
\end{tabular}
}
\vspace{-2mm}
\end{table*}

\section{Evaluation}
\label{sec:evaluation}

\urldef{\footurla}\url{https://figshare.com/articles/UI_logs/10269845}
\urldef{\footurlb}\url{https://github.com/volodymyrLeno/RPM}

We conducted a series of experiments to evaluate the performance of our approach when discovering different types of transformations. In addition, we tested the approach on a synthetically-recorded UI log that simulates a real-life use case, to verify its applicability on real-life scenarios.

\subsection{Experimental setup}
\label{sec:setup}

We built a dataset using the data transferring task presented in Section \ref{sec:running}. In this scenario, the students' contact information stored in an Excel spreadsheet has to be transferred into a Web form linked to a student management system. The task involves a number of data transformations of various types. We recorded 50 executions of such task by using the Action Logger tool \cite{DBLP:conf/bpm/LenoPRDM19}. We then preprocessed this log by segmenting it using the ``Submit button" action and applying the simplifier tool presented in Section \ref{sec:preprocessing}. The raw and preprocessed logs are publicly available.\footnote{\footurla}

Using this dataset, we evaluated the performance of three approaches: i) the baseline approach as per Section \ref{sec:naive} (Baseline), ii) the approach that involves target grouping as per Section \ref{sec:opt1} (Opt 1), and iii) the approach that uses both target grouping and grouping by input structure (Opt 1 + Opt 2). We measured the time required to discover the transformation program using all three approaches for different transformation types as well as for the entire UI log.\footnote{The experiments were conducted on a PC with Intel Core i5-5200U CPU 2.20 GHz and 16GB RAM, running Windows 10 as a host OS and a VM with Ubuntu 16.04 LTS (64-bit) with 8GB RAM and JVM 11 (4GB RAM).} We used the Foofah tool\footnote{\url{https://github.com/umich-dbgroup/foofah}} with a timeout of one hour. The tool we developed for this experiment, which implements the three approaches, is publicly available.\footnote{\footurlb}

\subsection{Results}
\label{sec:results}

\tablename~\ref{tab:transformationDiscoveryResults} describes how the three approaches perform on different types of transformations, while \tablename~\ref{tab:discoveryResultsUseCase} shows their computational efficiency on the entire UI log. The latter table also reports on the discovery quality of the approaches. The execution time is shown in seconds.

From Table~\ref{tab:transformationDiscoveryResults}, we can see that the baseline approach performs better than the two optimizations for very simple examples (three out of ten cases). This is due to the fact that it does not require any additional steps. However, in case of complex transformations its efficiency drops significantly. The second approach (Opt 1) discovered more transformations compared to the baseline. It outperforms the baseline in case of complex 1-N transformations as it decomposes the corresponding transformation into a set of small transformations of 1-1 type, and then discovers them separately. However, this approach was able to discover only a fraction of such transformations, as it could not deal with heterogeneous data (e.g., for cities/suburbs) and ambiguous output (e.g., for zip codes). In these cases, it took around half of hour to conclude that there is no transformation or that it is not possible to discover one. In contrast, the third approach (Opt 1 + Opt 2) discovered all data transformations within a reasonable time (up to 85 seconds). By clustering the patterns, this approach handles the problem of data heterogeneity, while by providing only one example from each equivalence class it deals with ambiguous output and speeds up the transformation discovery step. In some cases, however, it was slightly slower than the other two approaches, because of the high number of input patterns. Even if the transformation is simple, this approach discovers it for all patterns, thus requiring more time.

\begin{table}[h]
\caption{Transformation discovery results.\label{tab:discoveryResultsUseCase}}
\centering
\scalebox{0.75}{
\begin{tabular}{| c | c | c |}
\hline
\textbf{Approach} & \textbf{Execution time} & \textbf{Discovered transformations} \\ \hline
Baseline & 3742.669 & 0/9 \\
Opt 1 & 10551.536 & 5/9 \\
Opt 1 + Opt 2 & 130.854 & 9/9 \\ \hline
\end{tabular}
}
\end{table}

For the entire UI log, the baseline did not discover any transformation because it was too complex and involved many source and target elements, thus resulting in timeout. Since grouping by target allows us to split a large transformation problem into smaller problems so that each problem can be solved in isolation, Opt 1 was able to discover five out of nine data transformations. However, it took much longer than the baseline, because some small problems were too difficult to solve and required a considerable amount of time (nearly three hours). In contrast, Opt 2 discovered all nine transformations in around two minutes.

\section{Conclusion}
\label{sec:conclusion}

The paper presented an approach to analyze UI logs in order to discover routines that correspond to copying data from a spreadsheet or form to another. The key idea developed in the paper is that the effect of such routines is to transform a data structure consisting of the cells or fields read during the routine (input fields) into another data structure consisting of the modified cells or fields (output fields). Accordingly, the routine can be codified as a program that transforms a data structure into another. Such a transformation program can be discovered from the set of input-output examples induced by the routine's executions, via data transformation discovery techniques, such as Foofah. A direct application of Foofah leads to an overly large search space in practical scenarios. The paper shows that this search space can be pruned by analyzing the UI log in order to identify relations between input and output fields, and by grouping the examples according to the sequences of numeric and alphabetic tokens contained in the fields. An empirical evaluation demonstrates that the proposed technique (including its optimizations) can discover various types of transformations.


One of the limitations of the proposed technique is that it assumes that the output fields are entirely derived from (input) fields that are explicitly accessed (e.g., via copy operations) during the routine's executions. Hence, the technique cannot discover routines where a user visually reads from a field (without performing a ``copy'' operation on it) and writes what she sees into another field. 

Another limitation is that the technique assumes that the UI log consists of traces, such that each trace corresponds to a task execution. In some cases, a UI log may contain all the actions performed by a user during a recording session, which may last hours. Turning such a UI log into one that can be processed by our technique requires a segmentation technique capable of identifying the start and the end of each task execution. If there is a clear delimiter between one execution of a task and the next one (e.g., the user clicks on a ``submit'' button), this segmentation is trivial. But in the absence of such a delimiter, the segmentation of UI logs into traces may require more sophisticated techniques.

\textcolor{black}{There is also room for enhancing the proposed approach in terms of its inability to discover conditional behavior, where the condition depends on the value of an input field. Consider, for example, a routine that involves copying a delivery date. If the delivery country is USA, then the month comes before the day (MM/DD/YYYY), otherwise the day comes before the month. Here, the transformation depends on a condition of the form ``country = USA'', which the current approach is unable to discover (it only discovers conditions that depend on the patterns in an input field). Similarly, the approach cannot discover routines with (nested) loops, e.g., copying a purchase order that consists of multiple line items.}

Finally, other types of input patterns that can be detected could be considered. In the second of the proposed optimizations, we discover patterns in an input field consisting of alphabetic and numeric tokens with separators (e.g., phone numbers). However, the range of patterns that occur in practice is much wider. A possible extension is to combine the proposed technique with algorithms for discovering regular expressions and string transformations~\cite{DBLP:conf/edbt/JinCJKMH19}.
\textcolor{black}{Another related extension is to apply data extraction techniques, such as those embedded in FlashExtract~\cite{DBLP:conf/pldi/LeG14}, to discover copy-pasting routines where the input is a semi-structured document (e.g., a Web page) as opposed to a spreadsheet or a structured Web form.}


\medskip\noindent\textbf{Acknowledgments.} This research was funded by the Australian Research Council (DP180102839) and the  European Research Council (PIX project).

\bibliography{Leno}
\bibliographystyle{aaai}

\end{document}